\def\year{2022}\relax
\documentclass[letterpaper]{article} 
\usepackage{stylefile}  
\usepackage{times}  
\usepackage{helvet}  
\usepackage{courier}  
\usepackage[hyphens]{url}  
\usepackage{graphicx} 
\usepackage{natbib}  
\usepackage{caption} 
\DeclareCaptionStyle{ruled}{labelfont=normalfont,labelsep=colon,strut=off} 
\usepackage{algorithm}
\usepackage{algorithmic}
\usepackage{color}
\usepackage{amsfonts,amssymb}
\usepackage{amsmath}
\usepackage{microtype}
\usepackage{booktabs}
\usepackage{dashrule}
\usepackage{multirow}
\usepackage{arydshln}
\usepackage{newfloat}
\usepackage{listings}
\floatstyle{ruled}
\newfloat{listing}{tb}{lst}{}
\floatname{listing}{Listing}
\nocopyright
\title{Enhanced Speaker-aware \\Multi-party Multi-turn Dialogue Comprehension}
\author{
    Xinbei Ma\textsuperscript{\rm 1,2,3},
    Zhuosheng Zhang\textsuperscript{\rm 1,2,3},
	Hai Zhao\textsuperscript{\rm 1,2,3,\thanks{Corresponding author.}}\\
}
\affiliations{
    \textsuperscript{\rm 1}Department of Computer Science and Engineering, Shanghai Jiao Tong University\\
	\textsuperscript{\rm 2}Key Laboratory of Shanghai Education Commission for Intelligent Interaction\\
	and Cognitive Engineering, Shanghai Jiao Tong University, Shanghai, China\\
	\textsuperscript{\rm 3}MoE Key Lab of Artificial Intelligence, AI Institute, Shanghai Jiao Tong University, Shanghai, China\\
	{\tt
	sjtumaxb@sjtu.edu.cn, zhangzs@sjtu.edu.cn, zhaohai@cs.sjtu.edu.cn}
}

\usepackage{bibentry}

\begin{document}

\maketitle

\begin{abstract}
Multi-party multi-turn dialogue comprehension brings unprecedented challenges on handling the complicated scenarios from multiple speakers and criss-crossed discourse relationship among speaker-aware utterances. 
Most existing methods deal with dialogue contexts as plain texts and pay insufficient attention to the crucial speaker-aware clues.
In this work, we propose an enhanced speaker-aware model with masking attention and heterogeneous graph networks to comprehensively capture discourse clues from both sides of speaker property and speaker-aware relationships.
With such comprehensive speaker-aware modeling, experimental results show that our speaker-aware model helps achieves state-of-the-art performance on the benchmark dataset Molweni.  Case analysis shows that our model enhances the connections between utterances and their own speakers and captures the speaker-aware discourse relations, which are critical for dialogue modeling.
\end{abstract}

\section{Introduction}

Training models to understand dialogue contexts and answer questions has been shown even more challenging than common machine reading comprehension (MRC) tasks on plain text \cite{reddy2019coqa,choi2018quac}. In this paper, we focus on the challenging multi-party multi-turn dialogue MRC, whose given passage consists of multiple utterances announced by three or more speaker roles \cite{li2020molweni}. Compared to two-party dialogues \cite{Loweubuntu,Wudouban,Zhangedc}, multi-party multi-turn dialogues have much more complex scenarios: First, every speaker role has a different speaking manner and speaking purposes, which leads to a unique speaking style of each speaker role. Thus the speaker property of each utterance provides unique clues\cite{liumdfn,Gusabert}. Second, instead of speaking in rotation in two-party dialogues, the transition of speakers in multi-party dialogues is in a random order, breaking the continuity as that in common non-dialogue texts due to the presence of crossing dependencies which are commonplace in a multi-party chat. Third, there may be multiple dialogue threads tangling within one dialogue history happened between any two or more speakers, making interrelations between utterances much more flexible, rather than only existing in adjacent utterances. Thus, the multi-party dialogue appears discourse dependency relations between non-adjacent utterances, which leads up to a graphical discourse structure \cite{shi2019deep,li2020molweni}.

	\begin{table}[t]
		\begin{center}\small
			\begin{tabular}{|p{7.25cm}|}
				\hline
				\textbf{Context}: \\ 
				\textup{\textcolor[RGB]{16,83,154}{\textbf{\underline{benkong2}:} \textit{also i did a sudo chown -r and also got permission denied}} }
				
				\textup{\textcolor[RGB]{173,48,41}{\textbf{\underline{Dr\_Willis}:} \textit{swapfile drive ? you mean a swap partition ?}}}
				
				\textup{\textcolor[RGB]{16,83,154}{\textbf{\underline{benkong2}:} \textit{no a drive to share files with the rest of the network.} }}
			
				\textup{\textcolor[RGB]{173,48,41}{\textbf{\underline{Dr\_Willis}:} \textit{ok a 'share ' EMOJI is what ya mean . lol.. for samba ?}}}
			
				\textup{\textcolor[RGB]{11,138,17}{\textbf{\underline{NickGarvey}:} \textit{could you toss the commands and out put on pastebin ?}}} 
				
				\textup{\textcolor[RGB]{16,83,154}{\textbf{\underline{benkong2}:} \textit{error is : `` chown : changing ownership of FILEPATH operation not permitted ''}}}
			
				\textup{\textcolor[RGB]{227,109,38}{\textbf{\underline{smo}:} \textit{for vfat filesystems , the permissions are dictated by the mount options , not chmod}}}\\ 
				\hline
				\textbf{Question1}: \\
				\textit{What is the permission dictated by ?}\\   
				\textbf{Answer:}
				\textit{by the mount options}\\
				\textbf{Start Position:}
				\textit{471}\\
				\hline
				\textbf{Question2:}
				\textit{What system does Nick use?}
				
				\textbf{Answer:} $\langle\textup{no answer}\rangle$ \\
				\hline
			\end{tabular}
		\end{center}
		\caption{\label{table1} An example of multi-party multi-turn dialogue reading comprehension with one answerable question and one unanswerable question. Speaker names are underlined for highlight, and different colors indicate different speakers.}
	\end{table}

\begin{figure}[t]
		\centering
		\includegraphics[width=0.48\textwidth]{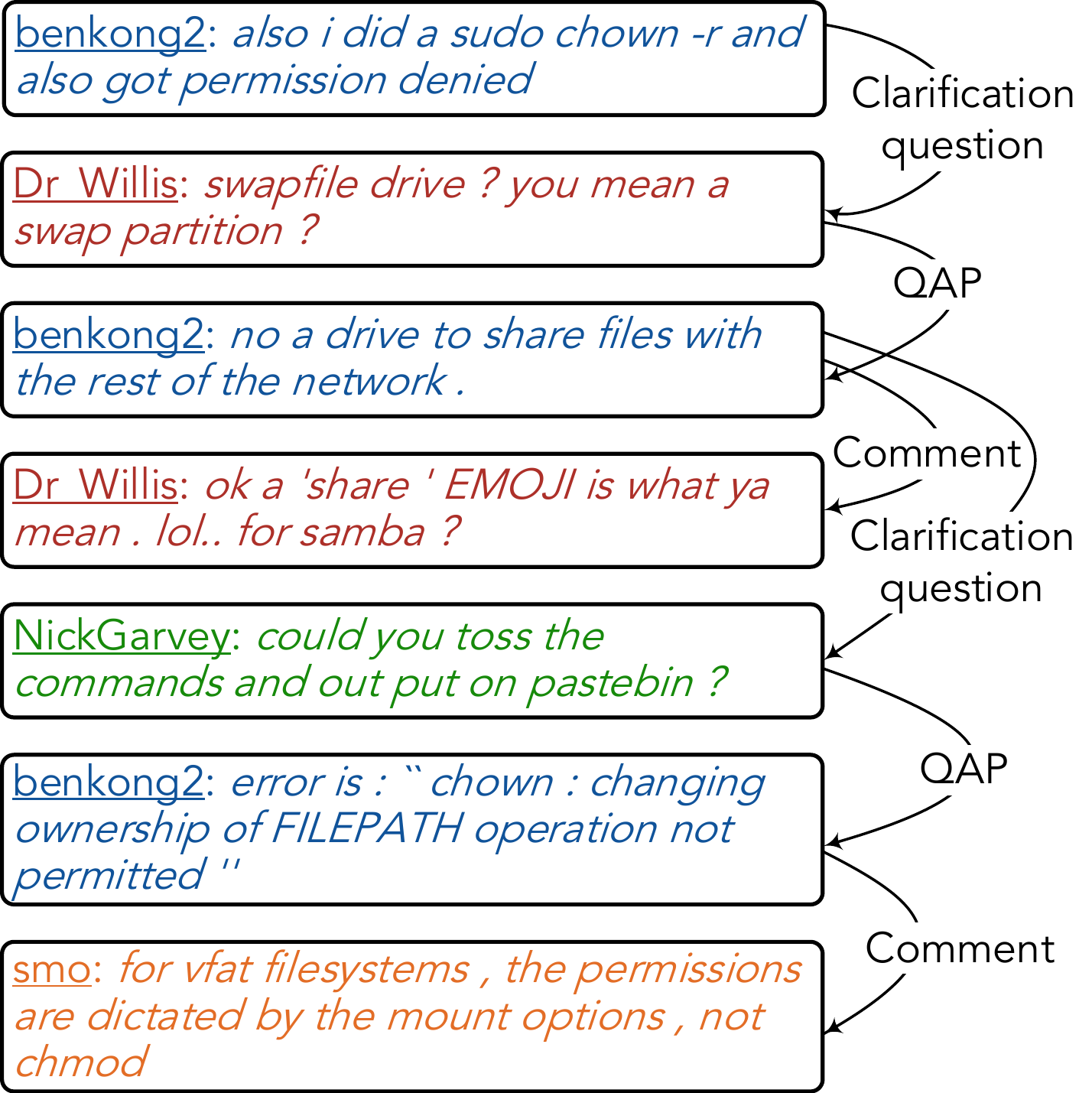}
		\caption{\label{discouse} Speaker-aware discourse structure of the example dialogue in Table \ref{table1}. }
\end{figure}

To demonstrate the challenge of the multi-party multi-turn dialogue MRC, we present an example in Table \ref{table1}, which is from the multi-party multi-turn dialogue benchmark dataset Molweni \cite{li2020molweni}. Figure \ref{discouse} depicts the corresponding speaker-aware discourse structure of the example dialogue, with different colors indicating different speakers. In this dialogue, there are four speakers, whose conversation develops as \emph{Dr\_Willis}, \emph{NickGarvey} and \emph{smo} help \emph{benkong2} with a system error. Along with the context, there are two relevant questions expected to be answered. An extractive span is given as an answer of question 1, while question 2 is unanswerable only based on this dialogue.

The mainstream work of machine reading comprehension on multiple multi-turn dialogues commonly adopts the pre-trained language model (PrLM) \cite{devlin2019bert} as an encoder to represent the dialogue contexts coarsely, taking the pairwise dialogue passage and question as a whole \cite{qu2019bert,Gusabert,li2020molweni}. Recent researches about modeling speaker-aware information for dialogue MRC proved to be effective \cite{Gusabert, liumdfn}. However, there are still unmatched attentions over key speaker role clues,

 $\bullet$ Attention paid on speaker role information is insufficient. The random and complicated speaker transition of multi-party dialogues needs to be disentangled and represented explicitly. 

 $\bullet$ Complex speaker role transition leads to sophisticated discourse structure but little attention is paid to structure and interrelations among the utterances, while discourse relationship among utterances may effectively embody speaker-aware clues from different perspectives.

In this work, we propose an enhanced speaker-aware model to comprehensively capture speaker-aware clues. In detail, 
to explicitly model the speaker role information, we employ extended disentanglement modules: 1) To capture the overall speaker-aware information in the entire dialogue, we operate masking-based multi-head attention method on each utterance on the basis of speaker roles. 2) We build two graph networks to model both annotated and unannotated discourse relations among utterances. These speaker-aware representions are fused and input a span-extraction layer to generate a reasonable answer to the question.

Experimental results on datasets show that the proposed strategy helps our model gain substantial performance improvements over strong baseline and achieve new state-of-the-art performance on Molweni \cite{li2020molweni} benchmark. 

\section{Background and related work}

\subsection{Dialogue Reading Comprehension}
Researches on dialogues MRC aim to teach machines to read dialogue contexts and make response \cite{reddy2019coqa,choi2018quac,sun2019dream,mutual}, whose common application is building intelligent human-computer interactive systems \cite{Chen2017survey,Shum2018,AliMe,zhu2018lingke}. Training machines to understand dialogue has been shown much more challenging than the common MRC as every utterance in dialogue has an additional property of speaker role, which breaks the continuity as that in common non-dialogue texts due to the presence of complex discourse dependencies which are caused by speaker role transitions  \cite{afantenos2015discourse,shi2019deep,li2020molweni}.  


Early studies mainly focus on the matching between the dialogue contexts and the questions \cite{huang2018flowqa,zhu2018sdnet}. As PrLMs prove to be useful as contextualized encoder with impressive performance, a general way is employing PrLMs to handle the whole input texts of a dialogue context and a question as a linear sequence of successive tokens, where contextualized information is captured through self-attention \cite{qu2019bert,liu2020hisbert, li2020molweni}. Such a way of modeling would be suboptimal to capture the high-level relationships between utterances in the dialogue history. 

To leverage speaker-aware information for better performance, \citet{Gusabert} proposed Speaker-aware BERT for two-party dialogue tasks by reorganizing utterances according to \emph{spoken-from speaker} and \emph{spoken-to speaker} and adding a speaker embedding at token representation stage. \citet{liumdfn} went further with speaker property, designing a decoupling and fusing network to enhance the turn order and speaker of each utterance. Both of them show that speaker property is helpful on dialogue MRC. However, existing studies mostly work on retrieval-based response selection task and on two-party datasets or those without speaker annotations, which drives us to make an attempt to extend to QA task on the multi-party scenario.

In this work, we focus on QA task of multi-party multi-turn dialogue MRC, which involves more than two speakers in a given dialogue passage \cite{li2020molweni} and expects an answer for each relevant question. Different from existing contributions of speaker-aware works, we regard discourse relations as a reflection of speaker transition information, thus leverage these complex relations to model speaker-aware information comprehensively.

\begin{figure*}[htb]
	\centering
	\includegraphics[width=0.97\textwidth]{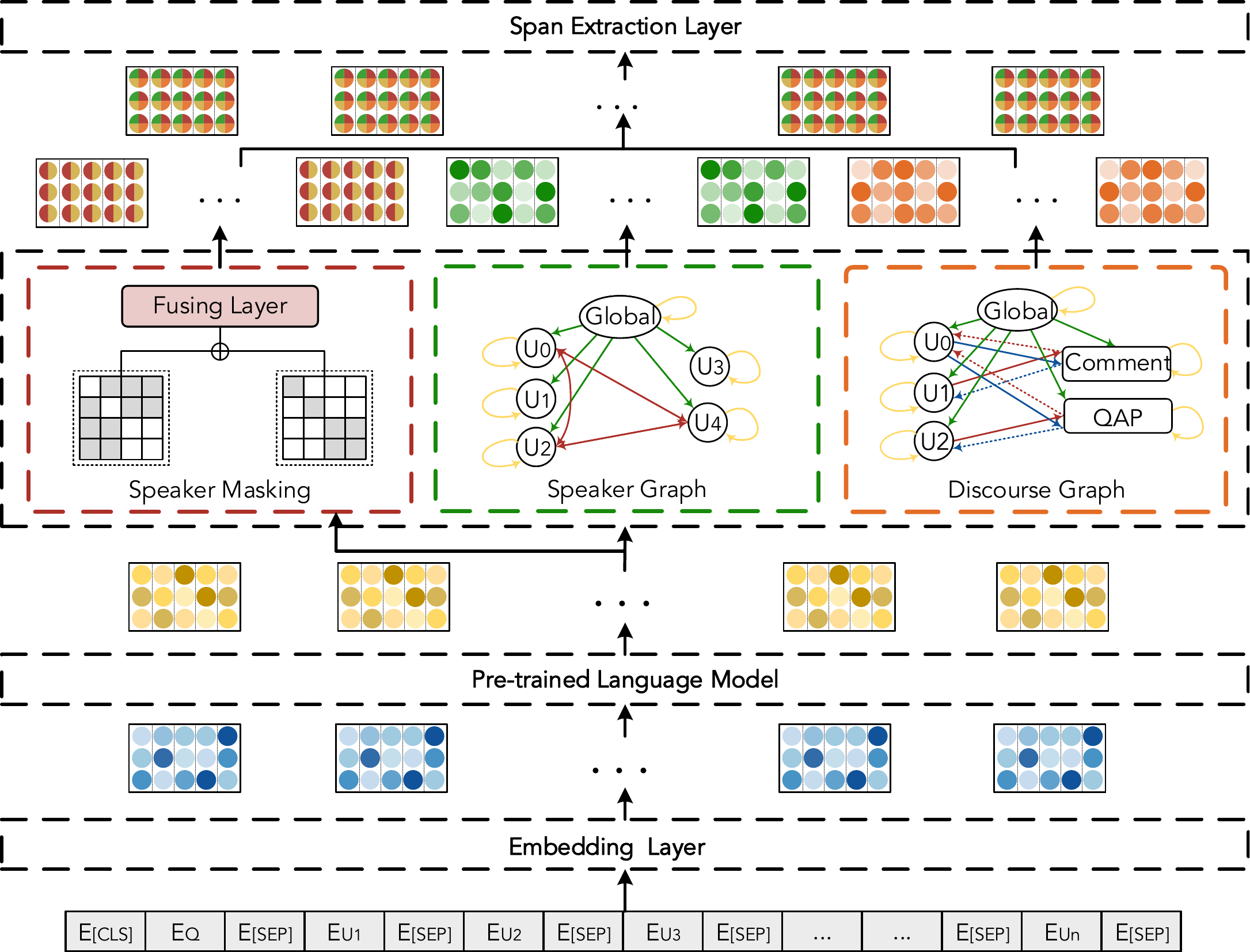}
	\caption{Overview of our model.}
	\label{overview}
\end{figure*}
\subsection{Discourse Structure Modeling}
Discourse parsing focuses on the discourse structure and relationships of texts, whose aim is to predict the relations between discourse units and to discover the discourse structure between those units. Discourse structure has shown benefits to a wide range of NLP tasks, including MRC on multi-party multi-turn dialogue \cite{asher2016discourse,xu-etal-2021-discovering,ouyang-etal-2021-dialogue,takanobu2018weakly,gao2020discern,jia2020multi}.

In addition to the concerned discourse parsing on dialogue-related NLP tasks, most existing studies on linguistics-motivated discourse parsing are based on two annotated datasets, Penn Discourse TreeBank (PDTB) \cite{prasad2008penn} or Rhetorical Structure Theory Discourse TreeBank (RST-DT) \cite{mann1988rhetorical}. 
PDTB focuses on shallow discourse relations but ignores the overall discourse structure \cite{qin2017adversarial,cai2017pair,bai2018deep,yang2018scidtb}.
In contrast, RST is constituency-based, where related adjacent discourse units are merged to form larger units recursively \cite{braud2017cross,wang2017two,yu2018transition,jotyetal-2015-codra,li2016discourse,liu2017learning}. However, RST only discovers the relations between neighbor discourse units, which is not suitable for our concerned multi-party dialogues. 

In this work, we use discourse parsing as an application-motivated technique for the dialogue MRC task. Our task relies on the dependency-based structures where dependency relations may appear between any two adjacent or non-adjacent utterances which may be presented by the same speaker \cite{shi2019deep,li2020molweni}. 


Compared to the existing works mentioned above, our work is distinguished because:
1) we leverage speaker-aware information comprehensively for better performance progress; 
2) we are one of the pioneers to model the speaker-aware discourse structure as graphs in dialogue MRC, to tackle the discourse tangle caused by speaker role transitions;
3) we firstly study general MRC task on multi-party multi-turn dialogue scenario with enhanced speaker-aware clues.
    

\section{Methodology}

Here, we present our enhanced speaker-aware model, as is shown in Figure \ref{overview}, which enhances speaker-aware information through three extended modules. Our model contains a PrLM for encoding, three modules for disentanglement of complicated speaker-aware information, namely, \emph{Speaker Masking}, \emph{Speaker Graph}, \emph{Discourse Graph}, and a span extraction layer for generating a final answer based on the fused representations. In this section, we will formulate the task and introduce every part of our model in detail.

\subsection{Task Formulation}
Supposing we conduct MRC on a multi-party multi-turn dialogue context $\mathbb{C}$, which consists of $n$ utterances and can be represented as $\mathbb{C}=\{U_1,U_2,\dots,U_n\}$. Each utterance $U_i$ consists of a name identity of the speaker and a sentence by the speaker, denoted by $U_i=\{S_i, W_i\}$, where the sequence $W_i$ can be denoted as a $l_i$-length sequence of words, $W_i=\{w_1,w_2,\dots,w_{l_i}\}$. According to this multi-party multi-turn dialogue context, a question $\mathbb{Q}$ is put forward, and for this question, the model is expected to find a span from the dialogue context as a correct answer, or make a decision that this question is impossible to answer only based on the provided dialogue context.
\subsection{Encoding}
In order to utilize a PrLM such as BERT as an encoder to obtain the contextualized representations, we firstly concatenate the dialogue context and a question in the form of \texttt{[CLS] question [SEP] context [SEP]}. For the convenience of dividing utterances, we insert \texttt{[SEP]} token between each pair of adjacent utterances. The concatenated sequence is fed into a PrLM, and the output of the PrLM is the initial contextualized representations for each token, denoted as $H \in \mathbb{R}^{L\times D}$, where $L$ denotes the input sequence length in tokens, $D$ denotes the dimension of hidden states.



\subsection{Speaker Masking}
Having obtained the output contextualized representations from a PrLM, we design a decoupling module to capture the speaker property of each utterance and represent the speaker transition information of the dialogue passage.

\begin{figure}[hbt]
		\centering
		\includegraphics[width=0.40\textwidth]{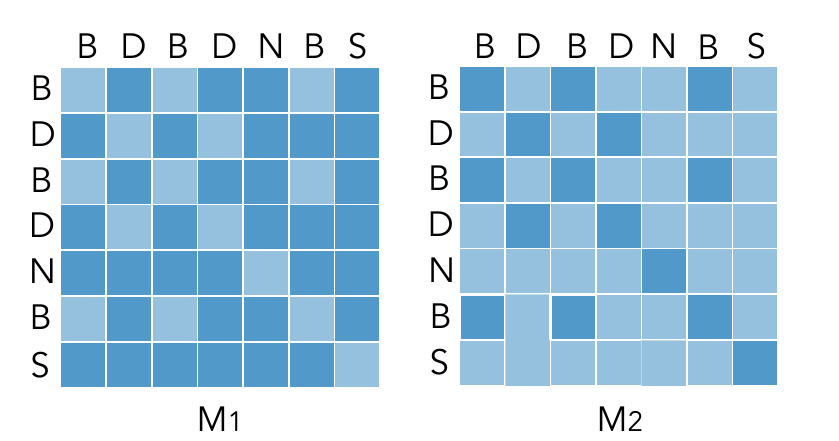}
		\caption{\label{mask} Speaker-aware Masking for the example shown in Table \ref{table1}.}
\end{figure}

We modify the mask-based Multi-Head Self-Attention mechanism proposed by \citet{liumdfn}, adapting it to multi-party dialouges. The mask-based MHSA is formulated as follows:
\begin{equation}\nonumber
\begin{split}
&\textit{A} \textup{(} \textit{Q, K, V, M} \textup{)} = \textup{softmax} \textup{(} \frac{\textit{QK}^{T}}{\sqrt\textit{d}_{k}} \textup{ $+$ } \textit{M} \textup{)} \textit{V} \textup{, }\\
&\textit{head}_{t} = \textit{A} \textup{(} \textit{HW}_{t}^{Q} \textit{, } \textit{HW}_{t}^{K} \textit{, } \textit{HW}_{t}^{V} \textit{, } \textit{M} \textup{), }\\
&\textit{MHSA} \textup{(} \textit{H, M} \textup{)} = \textup{[} \textit{head}_{1} \textit{, } \textit{head}_{2} \textit{, \dots ,} \textit{head}_{N} \textup{]} \textit{W}^{O} \textup{, }\\
\end{split}
\label{eq3—1}
\end{equation}

where $A$, $head_i$, $Q$, $K$, $V$, $M$ denote the attention, head, query, key, value and mask, $H$ denotes the original representations from PrLM, and $W_t^Q$, $W_t^K$, $W_t^V$, $W^O$ are parameters. Operator [$\cdot,\cdot$] denotes concatenation.
Instead of speaking in turn between two people, we have to explicitly identify the speaker of each utterance. In the implementation, we build a vector to label the speaker identity of each utterance, according to which, we mask utterances from the same speaker and utterances from different speakers. This step is denoted as:
\begin{equation}\nonumber
\setlength{\abovedisplayskip}{3pt}
\setlength{\belowdisplayskip}{3pt}
\begin{split}
\textit{M}_{1} \textit{[i, j]} &=
\begin{cases}
0,& \textit{S}_{i} \textup{$=$} \textit{S}_{j}\\
-\infty,& \text{otherwise,}\\
\end{cases}\\
\textit{M}_{2} \textit{[i, j]} &=
\begin{cases}
0,& \textit{S}_{i} \textup{$\neq$} \textit{S}_{j}\\
-\infty,& \text{otherwise,}\\
\end{cases}\\
\textit{Channel}_{1} &= \textit{MHSA} \textup{(} \textit{H} \textup{, } \textit{M}_{1} \textup{),}\\
\textit{Channel}_{2} &= \textit{MHSA} \textup{(} \textit{H} \text{, } \textit{M}_{2} \textup{),}
\end{split}
\label{eq2}
\end{equation}

where $S$ denotes the speaker identity, thus $M_1$ and $M_2$ denote masks of the same speaker and different speakers. ${Channel_1}$ contains the decoupled information of the same speaker while ${Channel_2}$ contains the decoupled information of the different speakers, as is shown in Figure \ref{mask}.

Finally, we fuse the information from ${Channel_1}$, ${Channel_2}$ and the original contextualized representation $H$ together, using the gate-based fusing method \cite{liumdfn}. The fusing method is formulated as:
\begin{equation}\nonumber
\setlength{\abovedisplayskip}{3pt}
\setlength{\belowdisplayskip}{3pt}
\begin{split}
\textit{E}_{1} &= \textup{ReLU} \textup{(} \textup{FC} \textup{([} \textit{H} \textup{, } \textit{C}_{1} \textup{, } \textit{H$-$C}_{1} \textup{, } \textit{H$\odot$C}_{1} \textup{]))},\\
\textit{E}_{2} &= \textup{ReLU} \textup{(} \textup{FC} \textup{([} \textit{H} \textup{, } \textit{C}_{2} \textup{, } \textit{H} \text{$-$} \textit{C}_{2} \textup{, } \textit{H$\odot$C}_{2} \textup{]))},\\
\textit{G} &= \textup{Sigmoid} \textup{(} \textup{FC} \textup{(([} \textit{E}_{1} \textup{, } \textit{E}_{2} \textup{])))},\\
\textit{H}_{C} &= \textit{G$\odot$C}_{1} \textup{$+$} \textup{(1$-$} \textit{G} \textup{)$\odot$} \textit{C}_{2},
\end{split}
\label{eq3}
\end{equation}
where $C_1$ and $C_2$ denote the shorthand of the two channels, and $\textup{FC}$ is shorthand of a fully-connected layer. Finally, we get the speaker-aware representations $H_C$, which is in the same size of the original contextualized representation $H$.

\subsection{Graph Modeling}
 Complicated transitions of speaker roles segment text into separated utterances and breaks the consistency of passage, thus results in intricate interrelations among utterances. We assume that these relations are reflection of speaker property, and will provide passage-level clues for MRC.
 
 We utilize the graph neural network to construct two heterogeneous graphs, called speaker graph and discourse graph, which are both in the form of relational graph convolutional network (R-GCN) following \citet{schlichtkrull2017modeling}. Speaker graph modeling relations of speaker property of each utterance. Discourse graph is built based on the speaker-aware discourse parsing relations, which are resulted from the complex non-adjacent dependencies caused by speaker transition and thus capture the latent speaker-aware information.
\subsubsection{Speaker Graph}
Since speaker property of each utterance impacts the dialogue development hugely, we build speaker graph to model relations of utterances based on speaker property. Specifically, we build an R-GCN to connect utterances from the same speaker, letting information exchanged among statements of one speaker, hoping to capture speaker manner. We denote the graph as $G_s=(V_s,E_s)$, where $V_s$ denotes the set of vertices and $E_s$ denotes the set of edges. First we add vertices $v_s^1,v_s^2,...,v_s^n$ to represent every single utterance and a special global vertice $v_d^{n+1}$ for context-level information, denoted as: 
\begin{equation}\nonumber
\setlength{\abovedisplayskip}{3pt}
\setlength{\belowdisplayskip}{3pt}
    V_s=(v_s^1,\dots,v_s^n, v_s^{n+1}), 
\end{equation}
where $n$ is the number of utterances. For each pair of utterances sharing the same speaker, we construct one edge and a reverse edge, which is denoted as $v_s^i \leftrightarrow v_s^j,S_i=S_j$. Finally, we construct a self-directed edge,  $v_d^i \rightarrow v_d^i$, for each vertice and we connect the global vertice to every other vertice, denoted as $v_d^{n+1} \rightarrow v_d^i,i \neq n+1$. 

\begin{figure}[htb]
		\centering
		\includegraphics[width=0.46\textwidth]{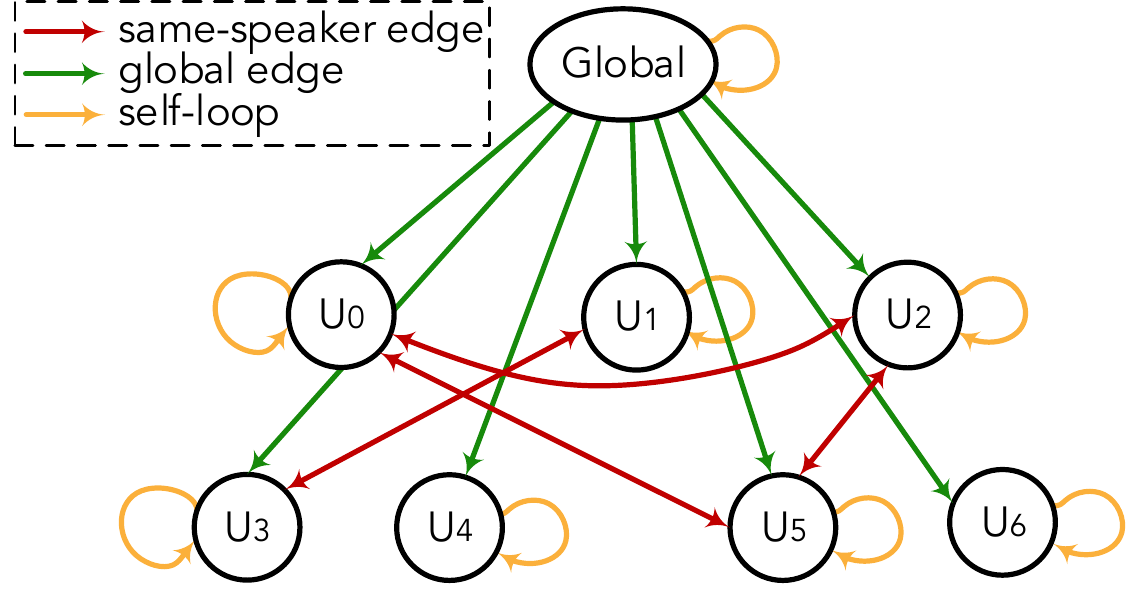}
		\caption{\label{sg} Speaker graph of the example dialogue in Table \ref{table1}.}
\end{figure}

\begin{figure}[hbt]
		\centering
		\includegraphics[width=0.48\textwidth]{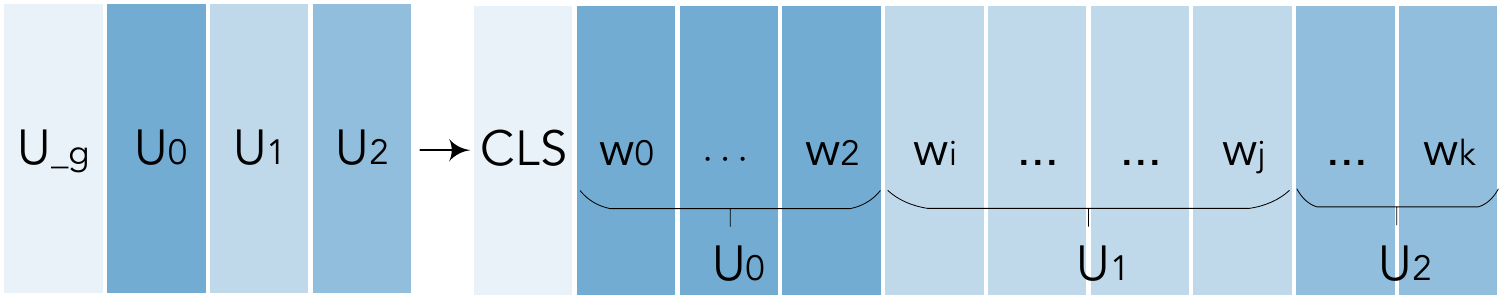}
		\caption{\label{extend1} Extension of output of speaker graph.}
\end{figure}

Figure \ref{sg} illustrates the graph structure of the example dialogue in Table \ref{table1}, with different colors for different kinds of edges.
The original representations of utterance vertice are the contextualized representations of \texttt{[SEP]} token extracted from $H$, and the original representations of the global vertice is formed by embedding. The information exchange process can be formulated as:
\begin{equation}\nonumber
\setlength{\abovedisplayskip}{3pt}
\setlength{\belowdisplayskip}{3pt}
\textit{h}_{i}^{(l+1)} = \text{$\sigma$($\sum_{r \in \mathbb{R}}\sum_{j\in N_i^r}\frac{1}{c_{i,r}}$} \textit{W}_{r}^{(l)} \textit{h}_{j}^{(l)} \textup{$+$} \textit{W}_{0}^{(l)} \textit{h}_{i}^{(l)} \textup{),}
\label{eq4}
\end{equation}
where $\mathbb{R}$ denotes the set of relations with other vertices. $N_i^r$ denotes the set of neighbours of vertice $v_i$, which are connected to $v_i$ through relation $r$, and $c_{i,r}$ is the element number of $N_i^r$ used for normalization. $W_r^{(l)}$ and $W_0^{(l)}$ are parameter matrices of layer $l$. $\sigma$ is activated function, which in our implementation is ReLU \cite{glorot2011deep, agarap2018deep}. After information exchange with neighbour nodes, we get the vectors of each utterance, containing speaker-aware interrelation information. After $L$ layers, we get $H_S^L \in \mathbb{R}^{(n+1)\times D}$ as the last-layer output of the graph. Based on the intuition that each token inside the same utterance shares the same speaker information, we expand $H_S^L$ to the same dimension of $H$ for later fusion, which is denoted as
$H_S \in \mathbb{R}^{L\times D}$. The extension is illustrated in Figure \ref{extend1}.

\subsubsection{Discourse Graph}
Discourse relations contain latent speaker-aware information. 
In parallel to the speaker graph, we build a graph according to the annotated discourse relations to connect relevant utterance pairs. The preprocessing includes two steps. First, we assign a label for every considered relation. Second, we simplify each relation in the form of \emph{(first utterance, second utterance, relation label)}.

\begin{figure}[hbt]
		\centering
		\includegraphics[width=0.46\textwidth]{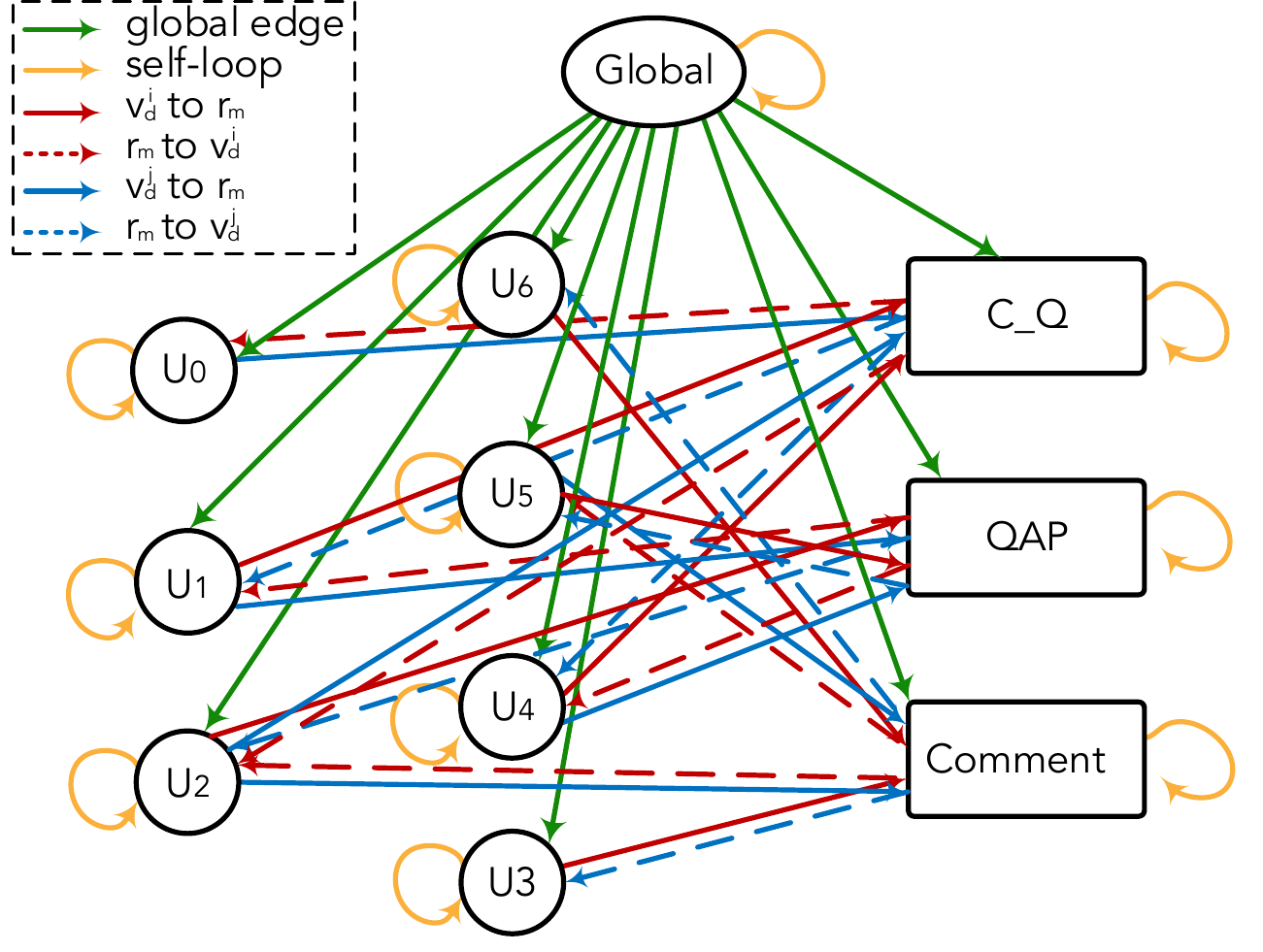}
		\caption{\label{graph} Discourse graph of the example dialogue in Table \ref{table1}.}
\end{figure}

Then the graph is constructed according to the simplified representations of relations. We denote the graph as $G_d=(V_d,E_d)$, where $V_d$ denotes the set of vertices and $E_d$ denotes the set of edges. Following kinds of vertices are constructed into the graph, utterance vertices for each utterances, relation vertices for each existing relations, and a global vertice to represent the dialogue-level information, which can be denoted as:
\begin{equation}\nonumber
\setlength{\abovedisplayskip}{3pt}
\setlength{\belowdisplayskip}{3pt}
    V_d=(v_d^1,\dots, v_d^n, v_d^{n+1}, v_d^{n+2},\dots, v_d^{n+{n_r}+1}), 
\end{equation}
where $n$ is the number of utterances and $n_r$ is the number of corresponding relations. In terms of $E_d$, for each relation (\emph{$v_d^i$, $v_d^j$, $r_m$}), we construct oriented edges $v_d^i\rightarrow r_m$ and $r_m \rightarrow v_d^j$, and also reverse oriented edges $r_m \rightarrow v_d^i$ and $v_d^j\rightarrow r_m$. As the same as speaker graph, we add a self-directed edge  $v_d^i\rightarrow v_d^i$ to every vertice and for each vertice except the global one, a global vertice-directed edge $v_d^{n+1}\rightarrow v_d^i, i\neq{n+1}$ is added. An example is shown in Figure \ref{graph}.
\begin{table}[t]
	\centering
	\setlength{\tabcolsep}{4pt}\small
	{
	\begin{tabular}{lcccc}  
	\toprule
	
	\multirow{2}{*}{\textbf{Model}} & \multicolumn{2}{c}{\textbf{Molweni}} & \multicolumn{2}{c}{\textbf{FriendsQA}} \\
		 &\textbf{EM} &\textbf{F1} &\textbf{EM} &\textbf{F1}  \\
        \midrule
    \multicolumn{5}{c}{\textit{BERT$_{base}$}}\\
    Public Baselines\cite{li2020molweni} & 45.3 &58.0 &45.2 & - \\
    \cdashline{1-5}[0.8pt/2pt]
    Our Baselines & 45.7  & 58.8 &45.6 & 61.0 \\
    +Speaker Embedding\cite{Gusabert}&47.9 &61.5 &45.0 &61.6 \\
    +MDFN\cite{liumdfn}&48.4 &62.4 &46.1 &62.9 \\
    +Our architecture&\textbf{49.7} &\textbf{64.4} &\textbf{47.0} &\textbf{63.0} \\
    \midrule
    \multicolumn{5}{c}{\textit{BERT$_{large}$}}\\
    Public Baselines\cite{li2020molweni} &51.8 &65.5 &- &-\\
    \cdashline{1-5}[0.8pt/2pt]
    Our Baselines & 52.0 & 65.6 &47.3 &63.3\\
    +Speaker Embedding\cite{Gusabert}&52.4 &65.7 &46.8 &63.3 \\
    +MDFN\cite{liumdfn}&51.7 &65.6 & 48.0 &63.0 \\
    +Our architecture&\textbf{52.9} &\textbf{66.9} &\textbf{49.0} &\textbf{64.0} \\
    \midrule
    \multicolumn{5}{c}{\textit{BERT$_{wwm}$}}\\
    Public Baselines\cite{li2020molweni} &54.7 &67.6 &- &-\\
    \cdashline{1-5}[0.8pt/2pt]
    Our Baselines & 53.9  & 67.5 &50.1 &66.2\\
    +Speaker Embedding\cite{Gusabert}&56.0 &68.3 & 49.2 &65.9 \\
    +MDFN\cite{liumdfn}&55.8 &68.7 &50.4 &66.2 \\
    +Our architecturel&\textbf{56.0} &\textbf{69.1} &\textbf{52.1} &\textbf{68.0} \\
    \midrule
    \multicolumn{5}{c}{\textit{ELECTRA}}\\
    Public Baselines\cite{li2020molweni} &- &- &- &-\\
    \cdashline{1-5}[0.8pt/2pt]
    Our Baselines & 57.3 & 70.4 &56.8 &74.0\\
    +Speaker Embedding\cite{Gusabert}&57.9 &57.9 &56.7 &74.0 \\
    +MDFN\cite{liumdfn}&57.9 &71.1 &57.8 &75.2 \\
    +Our architecture&\textbf{58.6} &\textbf{72.2} &\textbf{58.7} &\textbf{75.4}\\

	
	\bottomrule
	\end{tabular}
	}
	\caption{Experimental results on the test set of Molweni and FriendsQA. All results are from our inplementations except public baselines.}
	\label{t2}
\end{table}
\begin{table}
	\centering\small
	{\begin{tabular}{p{4cm}p{1.2cm}p{1.2cm}}
		\toprule
		\textbf{Model} & \textbf{EM} &\textbf{F1}\\ 
		\midrule
        \text{BERT$_{base}$}  & 45.3 &58.0 \\
        \quad +Speaker Masking &49.6  & 63.4\\
        \quad +Speaker Graph  &49.0  & 63.3\\
        \quad+Discourse Graph  &49.0  & 63.0\\
        +Our architecture  & \textbf{49.7} & \textbf{64.4} \\
        \midrule
        \text{BERT$_{large}$}  & 51.8 & 65.5 \\
        \quad +Speaker Masking &52.7  & 65.8\\
        \quad +Speaker Graph  &52.7  & 66.0\\
        \quad+Discourse Graph  &52.1  & 65.5\\
        +Our architecture  & \textbf{52.9} & \textbf{66.9} \\
        \midrule
        \text{BERT$_{wwm}$}  & 53.9 & 67.5 \\
        \quad +Speaker Masking &55.8  & 68.7\\
        \quad +Speaker Graph  &54.9 & 68.9\\
        \quad+Discourse Graph  &55.2  & 68.3\\
        +Our architecture  & \textbf{56.0} & \textbf{69.1} \\
        \midrule
        \text{ELECTRA}  & 57.3 & 70.4 \\
        \quad +Speaker Masking &57.9  & 71.0\\
        \quad +Speaker Graph  &57.6 & 72.1\\
        \quad+Discourse Graph  &58.4  & 71.8\\
        +Our architecture  & \textbf{58.6} & \textbf{72.2} \\
        \midrule
        \midrule
        \multicolumn{3}{c}{\textit{Ablation on BERT$_{base}$}}\\
		Our Model  & \textbf{49.7} & \textbf{64.4} \\
		\quad w/o Speaker Masking & 48.6 & 63.0\\
		\quad w/o Speaker Graph & 49.1 & 63.2\\
		\quad w/o Discourse Graph & 49.2 & 63.5\\

		\bottomrule
	\end{tabular}
	}
	\caption{Ablation study.}
	\label{t3}
\end{table}
Similar to the speaker graph, the original representations of utterance vertices are the contextualized representations of \texttt{[SEP]} token. The original representations of relation vertices and the global vertice are formed by embedding. Finally, we get the vectors of each utterance containing speaker-aware discourse structure information after the fusing of information from related vertices. The formulation of message-passing is the same as the speaker graph, where the set of relations $\mathbb{R}$ contains more kinds of relations as shown in Figure \ref{graph}. The output of the last-layer of discourse graph is denoted as  $H_G^L \in \mathbb{R}^{(n+n_r+1)\times D}$, we keep the vectors for utterances $H_G^L[0:n]$ and conduct the same extension as shown in Figure \ref{extend1}, then we get $H_G \in \mathbb{R}^{L\times D}$.

\subsection{Fusing}
Decoupled information from aforementioned three modules is fused to predict the answer. We concatenate $H_C$, $H_S$, $H_G$ and $H$ together to obtain the final speaker-enchanced contextualized representations: 
\begin{equation}\nonumber
\setlength{\abovedisplayskip}{3pt}
\setlength{\belowdisplayskip}{3pt}
\begin{split}
 \textup{P} &= \textup{[} \textit{H}_{C} \textit{,H}_{S} \textit{,H}_{G} \textit{,H} \textup{]}
\end{split}
\end{equation}

Following the standard process for span-based MRC \cite{devlin2019bert,glass2020span,zhang2021retro}, the representations are fed to a fully connected layer to calculate the probability distribution of the start and end positions of answer spans, and cross-entropy function is used as the training object to minimized.

\section{Experiments}
Our method is evaluated on multi-party multi-turn dialogue MRC benchmark, Molweni \cite{li2020molweni} and FriendsQA \cite{yang-choi-2019-friendsqa}.
\subsection{Datasets}
\subsubsection{Molweni}
Molweni \cite{li2020molweni} is a multi-party multi-turn dialogue dataset that derives from Ubuntu Chat Corpus \cite{Loweubuntu} which consists of 10,000 multi-party multi-turn dialogues context. On average, each dialogue context contains 8.82 utterances from 3.51 speaker roles. Following annotations are made on the raw dataset, making Molweni an ideal evaluation dataset for our research.
1) Answerable and unanswerable extractive questions according to dialogues.
2) Elementary discourse units (EDUs) on the utterance level, including the utterance and a speaker name.
3) Discourse relations for each dialogue passage, reflecting interrelations between utterances. 
\subsubsection{FriendsQA}
To verify the generality, we also evaluate our model on FriendsQA \cite{yang-choi-2019-friendsqa}, which is a challenging multi-party multi-turn dialogue dataset including 1,222 human-to-human conversations from the TV show \textit{Friends}. 10,610 answerable extractive questions are annotated. Discourse relations are annotated by using the tool of \citet{shi2019deep}.

\subsection{Baseline}
Following \citet{li2020molweni}, we use BERT as a naive baseline, where the contextualized output is used for span extraction directly. In addition, we compare our model with existing speaker-aware work \cite{liumdfn,Gusabert}. Since they work on response selection task on two-party scenario or datasets without explicit speaker annotations, we adjust and implement their ideas on QA task of the multi-party scenario.
We also apply BERT$_{large}$, and BERT$_{whol\,word\,masking}$ (BERT$_{wwm}$) and ELECTRA \cite{ClarkLLM20} as baselines, to see if the advance of our method still holds on top of the stronger PrLMs.

\subsection{Setup}

Our implementations are based on \textit{Transformers} Library \cite{wolf2020transformers}. Exact match (EM) and F1 score are the two metrics to measure performance. 
We fine-tune our model employing AdamW \cite{adamw} as the optimizer. The learning-rate is set to 3e-5, 5e-5, and 4e-6. In addition, the input sequence length is set to 348, which our inputs are truncated or padded to.


\subsection{Results}
Table \ref{t2} shows the results of our experiments. The experimental results show that our model outperforms all baselines and achieves SOTA on benchmark Molweni. We also see that our model helps effectively capture speaker role information and speaker-aware discourse structure information and then strengthens the ability of multi-party multi-turn MRC.


\section{Analysis}

\begin{figure}[htb]
		\centering
		\includegraphics[width=0.46\textwidth]{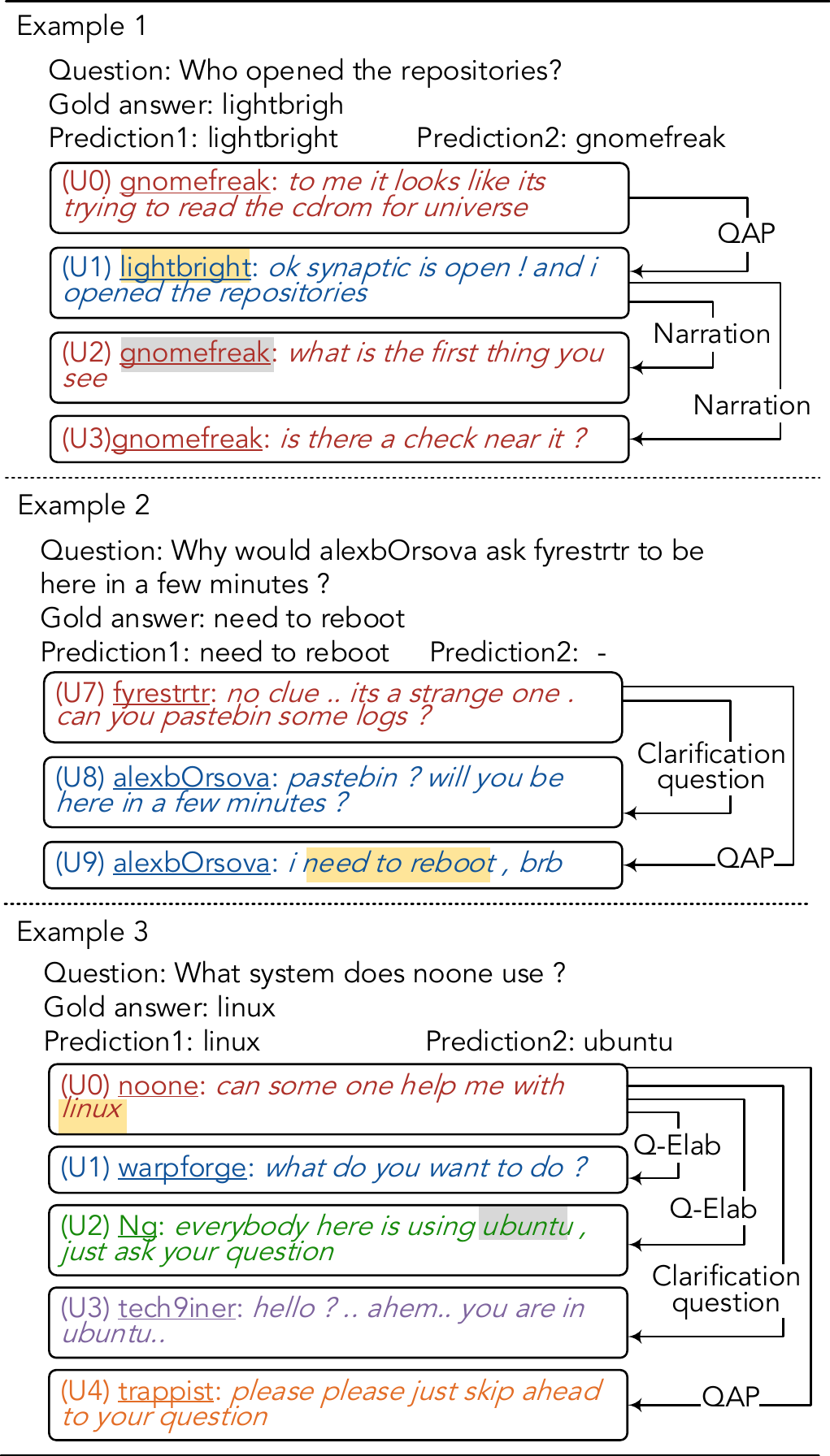}
		\caption{\label{case} Selected cases where baseline model fails (Prediction1) but our model gives gold answers (Prediction2). Related segments of dialogues are presented for illustration.}
\end{figure}

\subsection{Ablation study}
Since our speaker-aware information enhancing method includes three separate modules, we perform an ablation study to verify the contributions of our three speaker-aware modules. Respectively, we ablate each aforementioned modules and train them under the same hyper-parameters. As shown in Table \ref{t3}, experimental results indicate that each module plays an effective part in the whole model, and the Speaker Masking module contributes the most.


\subsection{Case Analysis}
To intuitively show how our model improves the ability of MRC on multi-party multi-turn dialogues, we present an analysis on the predictions from baseline (BERT$_{base}$) and predictions from our model to show how our speaker-aware enhancement strategies help fix wrong cases of baseline. 
We select examples of different types of questions and compare the predictions, as shown in Figure \ref{case}. 

In the first \emph{Who-type} case, the answer given by baseline model is \emph{gnomefreak}, which is the nearest speaker name to \emph{ opened the repositories}. While \emph{lightbright}, the answer given by our model is the gold answer, which is the speaker of the utterances containing the phrase \emph{opened the repositories}. Our model is able to fix this since we regard each utterance as an EDU and effectively model the speaker information. 

For the \emph{Why-type} question in case 2, the baseline model failed to find a plausible answer. However the Clarification-question relation and QAP relation among $u_7$ (from \emph{fyrestrtr}), $u_8$ (from \emph{alexbOrsova}) and $u_9$ (from \emph{alexbOrsova}) is very obvious, which are captured by our model.

In the third case, which is a \emph{What-type} case, the answer \emph{ubuntu} given by baseline model is reasonable already, based on $u_2$ which contains the key word \emph{use}. But our model gives the gold answer \emph{linux}, which is a more precise span from $u_0$, which is from \emph{noone}.

As these cases show, our model enhances the connections between utterances and their own speakers and captures the speaker-aware discourse relations, which helps to fix some wrong cases.

\section{Conclusion}
In this work, we study machine reading comprehension on multi-party multi-turn dialogues and propose an enhanced speaker-aware model to model speaker information comprehensively and firstly leverage discourse relation in dialogue MRC. 
Our model is evaluated on two multi-party multi-turn dialogue benchmarks, Molweni and FriendsQA. Experimental results show the superiority of our method compared to previous work. In addition, we analyze the contribution of each module by ablation study and present examples for intuitive illustration. Our work verifies that speaker roles and interrelations are significant characters of dialogue contexts. Our model takes advantage of enhancing the connections between utterances and their speakers and capturing the speaker-aware discourse relations.

\bibliography{bibfile}
\end{document}